\documentclass{article}

\usepackage{arxiv}
\usepackage{booktabs} 
\usepackage{graphicx}
\usepackage{tabularx}
\usepackage{adjustbox}
\usepackage{amsmath}
\usepackage{amssymb}
\usepackage{url}

\usepackage{enumitem}

\newlist{questions}{enumerate}{2}
\setlist[questions,1]{label=RQ\arabic*.,ref=RQ\arabic*}
\setlist[questions,2]{label=(\alph*),ref=\thequestionsi(\alph*)}
\title{AutoFITS: Automatic Feature Engineering for Irregular Time Series}

\author{Pedro Costa \\
	TeamViewer\\
	Porto, Portugal \\
	\texttt{pcosta.fcup@gmail.com} \\
	\And
	Vitor Cerqueira\\
	Dalhousie University\\
	Halifax, Canada\\
	\texttt{vitor.cerqueira@dal.ca} \\
	\And
	João Vinagre\\
	INESC TEC and University of Porto\\
	Porto, Portugal \\
	\texttt{jnsilva@inesctec.pt}
}

\date{} 

\renewcommand{\shorttitle}{AutoFITS}
\newcommand{\shortauthors}{P. Costa et al.}

\begin{document}

\maketitle

\begin{abstract}
A time series represents a set of observations collected over time. Typically, these observations are captured with a uniform sampling frequency (e.g. daily). When data points are observed in uneven time intervals the time series is referred to as irregular or intermittent. In such scenarios, the most common solution is to reconstruct the time series to make it regular, thus removing its intermittency.
We hypothesise that, in irregular time series, the time at which each observation is collected may be helpful to summarise the dynamics of the data and improve forecasting performance. We study this idea by developing a novel automatic feature engineering framework, which focuses on extracting information from this point of view, i.e., when each instance is collected.
We study how valuable this information is by integrating it in a time series forecasting workflow and investigate how it compares to or complements state-of-the-art methods for regular time series forecasting. In the end, we contribute by providing a novel framework that tackles feature engineering for time series from an angle previously vastly ignored. We show that our approach has the potential to further extract more information about time series that significantly improves forecasting performance.
\end{abstract}


\section{Introduction}

In a standard time series forecasting problem, the data is composed of two core elements: timestamps, which mark the point in time to which the observation refers, and a target variable, which is usually a numeric value (hence forecasting is commonly a regression problem). A regular time series is one where every consecutive observation is equally spaced in time with some temporal frequency \(f\) and many state-of-the-art methods assume a time series with such characteristics. 

Formally, if a regular time series has frequency \(f\), a set of timestamps \(T\), target variable \(Y\) and \(N\) observations, then, for \(\alpha \in \mathbb{N}\):

\[|t_j - t_i| = \alpha.f, \forall t_i,t_j \in T \]

In the above setting, the goal of forecasting is to predict a set of values \([y_{N+1},...,y_{N+h}]\), within a forecasting horizon \(h > 0\). In other words, $h$ is the length of time into the future for which we forecast the time series behaviour.

A problem arises when dealing with irregular time series, given that in these, timestamps are not evenly spaced. A typical approach is to resample the data, by aggregating observations into bins of equal time range. For instance, if we have \(T=[1,2,4,6]\) and \(Y=[1,2,3,4]\), resampling the data to a frequency 3, using a simple sum as the aggregator, would return \(T'=[1,4]\) and \(Y'=[3,10]\). From here, we can forecast from this time series using forecasting methods for regular time series.

However, the resampling approach may not be the best. The irregularity of the time series could itself contain relevant information. For instance, consider a time series consisting of restaurant reservations. Fewer reservations within a time period would imply less demand. Hence, a customer wanting to make a reservation may relax a bit more since there is probably a table available. By grouping the observations and tackling the problem as if it was a regular time series from the start, we are inevitably losing information that can be useful in the forecasting problem at hand. 

In this work, we explore the process of automatically extracting features from the irregularity present in some time series and show that, by exploiting the uneven temporal spacing between observations, we can achieve better forecasting performance. 

In summary, our contributions are the following:
\begin{itemize}
    \item An overview of the state-of-the-art for forecasting in irregular time series;
    
    \item An automatic feature engineering framework designed to extract information from the irregularity component of time series;
    
    \item A set of experiments which show the usefulness of the proposed approach.
    
\end{itemize}

On top of these, we contribute by developing a Python software package, in which our experiments are reproducible and extendable \footnote{\url{https://gitlab.com/blank.user.autofits/autofits}}.

We pose the following research question:

\begin{itemize}
    \item Can we improve a model's performance by extracting information about a time series irregularity? \label{rq:simple}
\end{itemize}

This research question is evaluated using two datasets that represent two scenarios with different complexity. The first dataset consists of a simple sequence of measurements over time from a single entity. The second dataset contains sequences by multiple distinct, yet related entities, each with its own measurements over time. In the latter case, the additional challenge is to build a forecasting model able to capture both local and global dynamics. 

In the remainder of this paper, we introduce related work in Section \ref{sec:related_work}, followed by the description of our proposal in Section \ref{sec:feature_extraction}. Our experimental setup is described in Section \ref{sec:experimental_setup}, and results are presented in Section \ref{sec:results}. We then conclude in Section \ref{sec:conclusions}.

\section{Related work}
\label{sec:related_work}
In this section, we overview the literature related to our work. We focus on two dimensions: feature extraction from time series, and forecasting with irregular time series. We list the main approaches to these problems and highlight our contributions.

\subsection{Time series feature extraction}

The information extracted from a time series may be used in multiple steps in a forecasting workflow. For this literature review, we look at works that extract features to improve on two processes: algorithm selection -- meta-learning -- and feature engineering.
model training (traditional feature engineering).

Regarding algorithm selection for forecasting, several works in the literature that use meta-learning to aid in this process, by first extracting meta-features from a time series and then using these to determine which method would be more appropriate to forecast an input series \cite{Prudencio2004, Lemke2010, Kang2017}. In essence, they extract descriptive features about the time series and/or the available methods for forecasting, and uses this information to select the best forecasting algorithm. 
Montero-Manso et al.~\cite{Montero-Manso2020} present an  ensemble approach where a meta-learner adjusts the relative weight of each base learner according to its predictive ability.

In \cite{Fulcher2014}, Fulcher and Jones introduce a method that extracts a set of thousands of time series features and constructs a feature-based classifier based on this set. Christ et al.~\cite{Christ2016} extend this work by implementing a highly parallel feature filtering. Fulcher and Jones \cite{Fulcher2017} attempt to refine \cite{Fulcher2014} and present \textit{hctsa}: a MATLAB-based implementation of their methodology. The method, however, generates thousands of features -- over 7.700, as stated in the original paper -- and in \cite{Lubba2019}, Lubba et al. show that it is possible to reduce the set of generated features from 4791 to just 22, with only an average decrease of 7\% in accuracy scores.

In \cite{cerqueira2020vest}, Cerqueira et al. propose VEST, a framework that attempts to produce an optimal time series representation for forecasting tasks. This is achieved in 3 steps: a transformation step - where the time series is transformed into several different representations; a summarising step - where each representation is summarised using statistics and a selection phase - where feature selection is applied to reduce the high number of features generated in the previous steps. VEST has a set of 8 transformation functions available and 32 summarising functions, which results in a set of 256 extracted features for feature selection.

All of these works assume a regular time series is being handled. Our approach aims to help forecast irregular time series and complement regular forecasting methods.

\subsection{Irregular time series forecasting}

Concerning forecasting time series with unevenly spaced observations, most work done in this area attempts to translate this into a regular time series forecasting problem and proceed from there - time series reconstruction. Heck et al.~\cite{heck1985period} distinguish between two families of algorithms to achieve this: Fourier transformation-based and non-parametric techniques derived from the \(\theta\)-criterion inspired by Lafler and Kinman \cite{lafler1965rr}. On the same line, Adorf \cite{adorf1995interpolation} presents several interpolation techniques aimed at irregularly sampled time series. It not only discusses the techniques mentioned before, but also talks about several other such as Matrix Inversion~\cite{kuhn1982recovering} techniques, Least-square Estimation~\cite{barning1963numerical, ferraz1981estimation}, Autoregressive Maximum-Entropy Interpolation~\cite{fahlman1982new, brown1990technique}, and Polynomial Interpolation~\cite{groth1975probability}.  

However, our work aims to create a feature engineering method capable of extracting information about a time series' irregularity. Croston~\cite{croston1972forecasting} introduces the Croston method: a forecasting strategy for products with intermittent demand using only historical data about the demand/sales of a product. We consider this method to be towards ``irregular forecasting'' as it attempts to extract information from the previous time periods where there was demand (which are inherently irregular), almost ignoring days with zero demand. A lot of adaptations of the Croston method have surged in the literature such as \cite{SYNTETOS2005303}, \cite{PRESTWICH2014928} and \cite{TEUNTER2011606}. In \cite{WILLEMAIN2004375}, Willemain et al. propose a method based on bootstrapping -- a statistical technique involving random sampling with replacement -- on previous observations of non-zero demand. The authors claim to have significantly improved over the Croston method, although this claim has been formally challenged later \cite{gardner}. 

Temporal aggregation is a technique that attempts to forecast an irregular time series by grouping observations into time periods. Nikolopoulos et al. \cite{Nikolopoulos} introduce an Aggregate - Disaggregate Intermittent Demand Approach (ADIDA) to forecasting. Briefly explained, after aggregating the observations, any standard forecasting method can be applied such as simple exponential smoothing. The forecast should then be disaggregated into time periods of the original size, using some sort of heuristic. Our work adopts the temporal aggregation strategy and creates a framework focused on information extraction about the irregularity of the series, by extracting features from the original observations that fall within the new time intervals. We set ourselves apart from methods created towards solving intermittent demand problems by allowing our framework to be able to be used with any irregular time series. 
We provide flexibility to the user to be able to extract information in any sort of domain effectively and intuitively and, despite the main focus being feature engineering and information extraction about the series' irregularity, we integrate our features into a simple time series forecasting workflow that takes advantage of such features. With that being said, we do not intend to compete with state-of-the-art forecasting methods, but rather complement and aid them by extracting information from an unusual viewpoint.

\section{Feature extraction from irregular time series}
\label{sec:feature_extraction}

AutoFITS extracts features from the existing irregularity present in the timestamps. These features encompass a wide range of statistics that may help highlight some underlying irregular structure to the time series and summarise the dynamics of the data. 
Our hypothesis is that this process may improve forecasting performance.

While preserving the original observations -- to keep information about the irregularity, we split the feature extraction process into 3 steps:
\begin{enumerate}
    \item Resampling the time series to a regular frequency using some aggregation and imputation strategy. Aggregation refers to how we group the observations (a ``mean'' aggregation strategy implies that observations that fall between a time interval will be averaged), and imputation refers to how we fill missing values, i.e, time intervals without any corresponding observations in the original data.
    \item Creating the time-delay embedding representation based on Taken's theorem~\cite{takens}, using the resampled data.
    \item For each time-delay embedding observation, extract features from the original observations that fall within that time frame, as well as features about the embeddings themselves.
\end{enumerate}

Summarising, for an irregular time series \(\mathcal{T} = [(t^0_1,y^0_1),...,(t^0_n,y^0_n)]\), AutoFITS first creates a regular time series \(\mathcal{T}_{reg}\) of size \(k\) through resampling -- temporal aggregation -- \(\mathcal{T}\). Then, \(\mathcal{T}_{reg}\) is transformed into a time-delay embedding representation of size \(l\):

\begin{equation*}
\mathcal{T}_{reg} =
    \left[
    \begin{array}{cccc|c}
        y_1 & y_2  & ... & y_{l} & y_{l+1}\\
        y_2 & y_3  & ... & y_{l+1} & y_{l+2}\\
        \vdots & \vdots  & \vdots & \vdots & \vdots\\
        y_i & y_{i+1}  & ... & y_{i+l-1} & y_{i+l}\\
        \vdots & \vdots  & \vdots & \vdots & \vdots\\
        y_{k-l} & y_{k-l+1}  & ... & y_{k-1} & y_{k}\\
    \end{array}
    \right]
\end{equation*}
\\
For each embedding \([y_i , y_{i+1}  , ... , y_{i+l-1}]\), the corresponding timestamps are \([t_i , t_{i+1}  , ... , t_{i+l-1}]\). As such, it represents a recent past of \(y_{i+l}\), i.e., the time interval \((t_i, t_{i+l-1})\). Original observations with timestamps within this interval are used to extract irregular features.

In Table~\ref{tab:features}, we display all features created by AutoFITS. We describe them based on whether they are calculated from the resampled data or from the original data -- column RD --, and whether they extract information from the time series' irregularity -- column I.

We extract information that directly models the frequency of observations by calculating statistics about the temporal difference between consecutive observations in the original data -- T\_DIF\_STATS. Features T\_Y\_AVG\_MUL and T\_Y\_AVG\_DIF\_MUL also indirectly serve this purpose as they calculate some tendency in the evolution of the timestamps and the target variable by calculating the average over the application of mathematical operations over the original features. Such type of operations have been shown to extract additional information from regular features ~\cite{Markovitch2002, Piramuthu2009, Dor2012}. 

To bring out information about the time interval where observations were recorded and the scope of the observed values, we create features such as 2D\_SPACE\_AREA, MIN\_MAX\_T\_DIF and MIN\_MAX\_T\_DIF\_F.

Another set of features is based on the level of irregularity in the time series. We create features such as REL\_DISP\_T, obtained by measuring the relative dispersion using the coefficient of variation, given by \(r_{disp} = \frac{std\_dev}{mean}\) and the entropy of the timestamps ENTROPY\_T. MISSING\_T\_COUNT may also provide some hints about this distribution by emphasising \emph{slow periods}, i.e., intervals in the resampled time series that do not have any corresponding observations on the original data. 

To model the recent past and attempt to explain the evolution in the target variable we create features such as MOV\_AVG -- the basis for moving average models --, and REG\_MOD as a way of assessing how easily can two regression algorithms model the recent past. 

Finally, we create some features based on both the resampled and the original data, a task made easy by our implementation. For example, ENTROPY\_Y or REL\_DISP\_Y are features originally created to model irregularity when applied using the timestamps, but that can also be used in regular data to extract other potentially useful features.

\begin{table*}[htb]
        \caption[Summary of features created by AutoFITS]{Summary of features created by AutoFITS.}
        \begin{tabularx}{\textwidth}{lXlc}
            \toprule
            \textbf{Name} &   \textbf{Description} & \textbf{I}  &  \textbf{\shortstack{RD}}\\
            \midrule
             REL\_DISP\_T & Relative dispersion of the timestamps.  & Yes & Both\\

            T\_Y\_AVG\_MUL & Average of the multiplication between the timestamps and the embeddings.  & Yes & Both\\
            
            T\_Y\_AVG\_DIF\_MUL & Average of the multiplication between the timestamps' time difference and the embeddings difference between consecutive observations.  & Yes & Both\\
            
            2D\_SPACE\_AREA & Time difference between the oldest and newest timestamp multiplied by the difference between the maximum and minimum embedding. & Yes & Both\\
            
            MISSING\_T\_COUNT & Number of timestamps missing, i.e, observations in the resampled data that have no corresponding original observations. & Yes & Yes\\
            
            T\_DIF\_STATS &   Statistics about time differences between consecutive timestamps, namely arithmetic mean, standard deviation, variance, sum, median, interquartile range (IQR), minimum, maximum and relative dispersion.  & Yes & No\\
            
            MIN\_MAX\_T\_DIF & Time difference between the oldest and newest timestamp. & Yes & No\\
            
            MIN\_MAX\_T\_DIF\_F & Time difference between the oldest and newest timestamp divided by the resampling frequency. & Yes & No\\
            
            ENTROPY\_T & Entropy of the timestamps. & Yes & No\\
            
            ENTROPY\_Y & Entropy of the embeddings. & No & Both\\
            
            REL\_DISP\_Y & Relative dispersion of the embeddings.  & No & Both\\
            
            MOV\_AVG & A moving average over the embeddings. The default window size is 3, but it can be configured. & No & Yes\\ 
            
            REG\_MOD & Results of applying a LASSO~\cite{lasso} and LinearRegression model to the available data. \(Y\) is the embeddings and \(X\) is the rest of the original features. We train a model on \(n-1\) observations and attempt to predict \(y_n\). The errors and the prediction itself result in the new features. & No & Yes\\
            
            \bottomrule
        \end{tabularx}
        \label{tab:features}
\end{table*}

\section{Experimental setup}
\label{sec:experimental_setup}

For our experiments, we integrate AutoFITS in a standard time series forecasting workflow. Our code is openly available at \footnote{\url{https://gitlab.com/blank.user.autofits/autofits}}. 

We first apply standard data pre-processing, such as missing value imputation, and splitting date features -- except the timestamps -- into their components -- year, month, day, etc. After these operations, the feature extraction process ensues. When dealing with multiple entities in the time series, the feature construction process is run separately for each entity's observations. It follows a disaggregate-aggregate strategy, where each entity is processed separately. Then, the resulting datasets are merged to create the new time series with multiple entities. 

We experiment with a LASSO and a RandomForest learning algorithm.
To validate our models, we implement the holdout method with 10\% for the test set size. In setting with multiple entities, we extract the last 10\% observations for each entity in the resulting dataset. 

The final step in the workflow is forecasting. We aim to be able to forecast the next value in the time series. This value is dependent on the set frequency, as it is representative of a value recorded at a time interval of duration \(f\). For example, consider a dataset with observations recorded from January 1 until January 10. We use \emph{1 day} as frequency and the arithmetic sum as an aggregator. In this setting, the framework would forecast a value for \(t =\) January 11, representing the sum of the target variable from January 11 to January 12. For instance, If we are dealing with visitors to a restaurant, this value would signify the total number of visitors during that day. In case we are dealing with multiple entities, the forecasting will be constituted by multiple forecasts, one for each entity, provided each entity has enough observations, otherwise it is ignored. 

\section{Results}
\label{sec:results}

\subsection{Evaluation}

To assess the added value of AutoFITS as an automated feature engineering framework, we create 2 models that work similarly to it, except in the feature creation stage: 
\begin{itemize}
    \item BaselineFITS: Simple forecasting baseline without feature creation. It is used as a way to compare performance with and without feature extraction.
    \item AutoFV: Merges features created by AutoFITS and VEST~\cite{cerqueira2020vest}. Used to evaluate how AutoFITS may complement standard feature engineering methods for regular time series.
\end{itemize}

Considering VEST was built for handling univariate time series, we adapt it VEST so it is able to handle multiple entities in a dataset. It follows the same logic as AutoFITS as it works by running VEST in parallel for the observations for each entity and then merging the results in a singular dataset.

In terms of evaluating the performance of the models, we use two regression metrics: Mean Absolute Error (MAE) and \(R^2\). 

\subsection{Datasets}

We use 2 datasets in our experiments, with each one representing a problem setting we intend to solve with AutoFITS.

This first dataset is the Vostok Ice Core dataset, from the paper by Petit et al.~\cite{vostok}. It refers to historical data recorded in Vostok, East Antarctica. At this remote location, it was possible, by studying the composition of the ice core composition, to obtain an accurate historical record about the atmosphere throughout thousands of years, since air bubbles get trapped in the ice. This dataset contains two features: 

\begin{itemize}
    \item \texttt{age (yrs bp)}: The \emph{timestamp}, presented in years BP (it is common practice to use January \(1^{st}\) 1950 as the epoch of the age scale). Marks the time at which the observation refers to.
    \item \texttt{co2}: The target variable. Measures the \(CO_2\) concentration levels in parts-per-million (\emph{ppm}) at a certain point in time.
\end{itemize}

The second dataset is the Recruit Restaurant Visitors dataset and was originally made public in a Kaggle competition~\footnote{https://www.kaggle.com/c/recruit-restaurant-visitor-forecasting/overview}. It collects data concerning visitors and reservations in a variety of restaurants to create a predictive model capable of forecasting the total number of visitors to a restaurant for future dates. Although this dataset is a relational dataset with multiple tables, we experiment using only the table related to reservations made to the restaurants, containing only 4 attributes:

\begin{itemize}
    \item \texttt{air\_store\_id}: Restaurant identifier (entity id).
    \item \texttt{visit\_datetime}: When is the reservation (timestamp).
    \item \texttt{reserve\_datetime}: When was the reservation made. This attribute is discarded.
    \item \texttt{reserve\_visitors}: For how many people is the reservation (target).
\end{itemize}

There are two main reasons behind this:
\begin{enumerate}
    \item We want to assess the value of the features generated by our approach when there is minimal information about the problem setting.
    \item This table represents a time series where sample times vary greatly. The dataset contains tables such as \texttt{Air Visit Date}, where sample times also vary, but in a more controlled way since missing observations are related to weekends, and other rare days when the restaurant is closed. Hence, the difference between observations is usually under 3 days. Considering that \texttt{Air Reserve} contains information about visitors with a reservation, the data is subjected to more factors that make it so it varies quite sporadically. 
\end{enumerate}

\subsection{Workflow parameters}

To train our models, we experiment using a LASSO and a RandomForest learner. For resampling the data, we set the parameters according to the dataset at hand: for the Recruit Restaurant Visitors dataset we aggregate observations using sum to represent the total amount of reserve visitors, impute zeros, since missing observations mean there were 0 reservations made, and test with frequencies ranging from 1 to 31 days, in steps of 1; for the Vostok Ice Core dataset, considering that the target variable represents \(CO_2\) concentration values in the atmosphere, we resample the data using a mean resampling strategy, impute the last valid observed value -- forward-fill -- and, considering the timestamps are represented in \emph{Years BP}, experiment with frequencies ranging between 250 and 4500 years, in steps of 250.

The last parameter to set is the lag size \(l\) that defines the amplitude of the recent past used in the time-delay embedding representation. For the Vostok Ice Core dataset, we use \(l=7\), yet, with the Recruit Restaurant Visitors dataset, we change \(l\) as we increase \(f\). Hence, for \(1 \leq f < 14\), we use \(l=10\); for \(14 \leq f < 21\), we use \(l=7\); for \(21 \leq f < 28\), we use \(l=3\) and ; for \(28 \leq f\), we use \(l=2\). This is mostly due to the fact that, as we increase \(l\), we are increasing the \emph{demand} for more observations to make the time-delay embedding representation. We do not impute values beyond the timestamps in the dataset, so if in the case of multiple entities, an entity does not have enough observations to create at least one time-delay embedding, that entity is discarded. Values for \(l\) were therefore chosen to minimise the trade-off in model performance and discarded entities.

\subsection{Results}

\subsubsection{Single entity}

\begin{figure}[htb]
    \centering
    \begin{minipage}{.48\textwidth}
    \includegraphics[width=\linewidth]{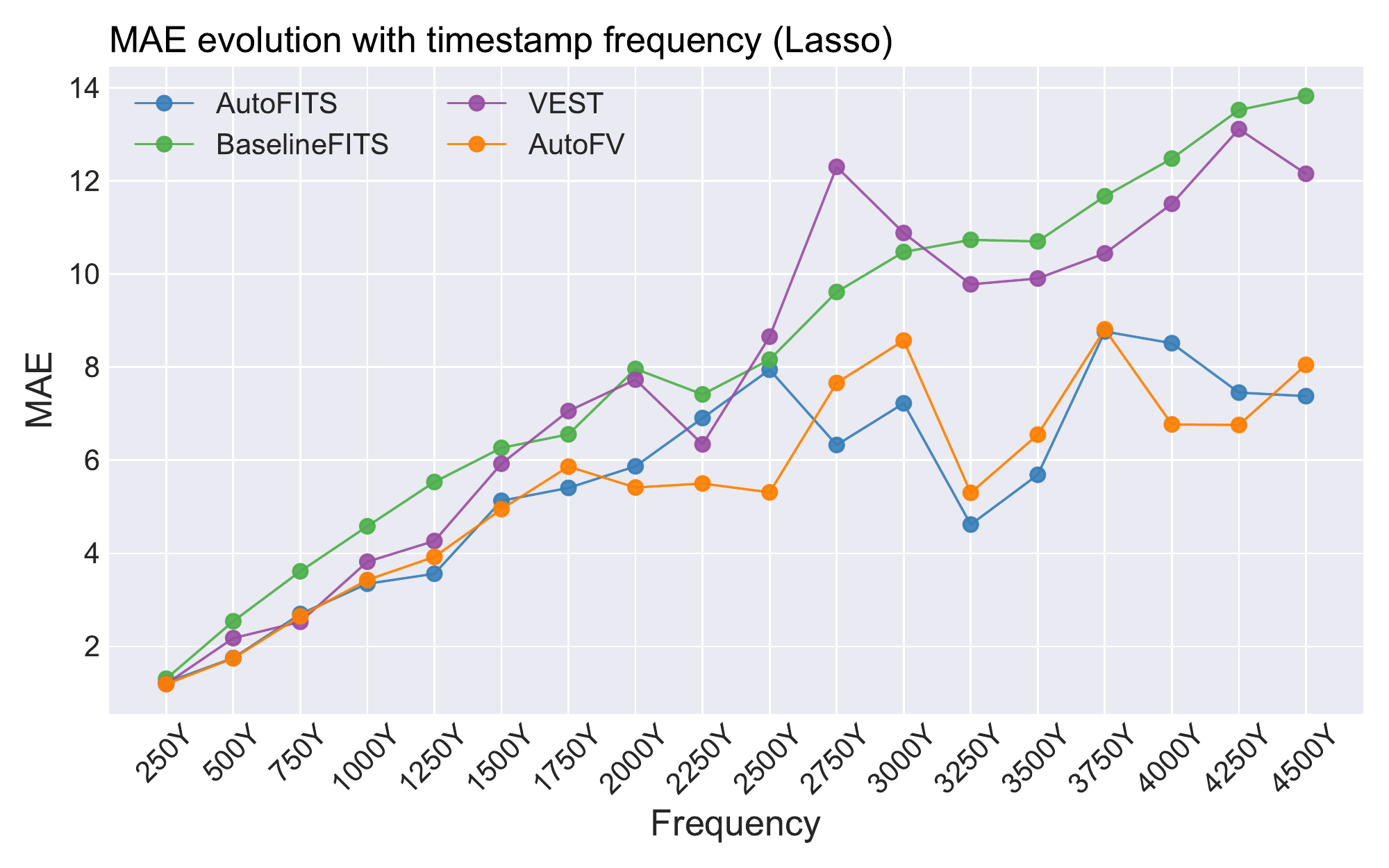}
    \caption{Vostok Ice Core dataset: MAE with a LASSO learner.}
    \label{fig:mae_vostok_lasso}
    \end{minipage}%
    \hspace{0.5cm}
    \begin{minipage}{.48\textwidth}
    \includegraphics[width=\linewidth]{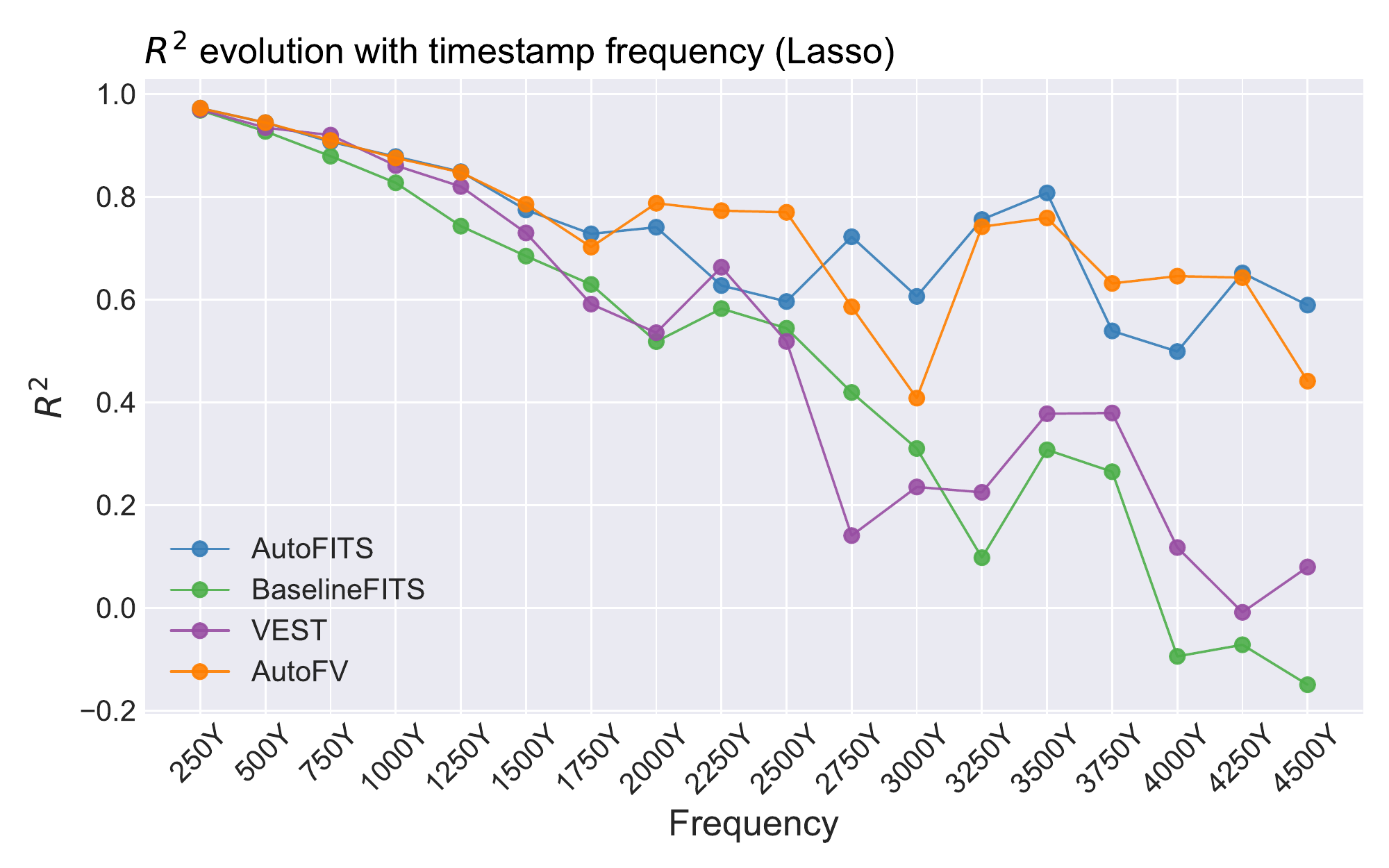}
    \caption{Vostok Ice Core dataset: \(R^2\) with a LASSO learner.}
    \label{fig:r2_vostok_lasso}
    \end{minipage}
\end{figure}

In Figure~\ref{fig:mae_vostok_lasso}, we plot the Mean Absolute Error evolution, as we increase the frequency for the Vostok Ice Core dataset. There is a clear ``line'' between the performance of models that take advantage of the series' irregularity - AutoFITS and AutoFV - and models that do not - BaselineFITS and VEST. In total, we ran 4 models for a set of 18 frequencies. From those experiments, AutoFITS and AutoFV had the lowest MAE in 9 (50\%) and 8 (44\%) cases respectively, while VEST did so in 1 case (11\%). It is worth noting how AutoFV performed the best in a significant amount of cases, bringing further credibility to the idea that AutoFITS can help improve standard forecasting methods.

In the \(R^2\) plot (Figure~\ref{fig:r2_vostok_lasso}), it becomes even more evident how much better AutoFITS and AutoFV perform against VEST and BaselineFITS. In almost all cases, we get \(R^2 \geq 0.6\) for both models that learn from irregularity. Moreso, BaselineFITS and VEST tend to have it \(R^2\) decrease rather quickly as the frequency increase, yet both AutoFITS and AutoFV seem to somewhat keep a good \(R^2\) as the frequency increases. Moreover, we further reinforce the theory that merging features created by VEST and AutoFITS adds value to a forecasting model, due to AutoFV also having the largest \(R^2\) in 8 cases.

\begin{figure}[htb]
    \centering
    \begin{minipage}{.48\textwidth}
    \includegraphics[width=\linewidth]{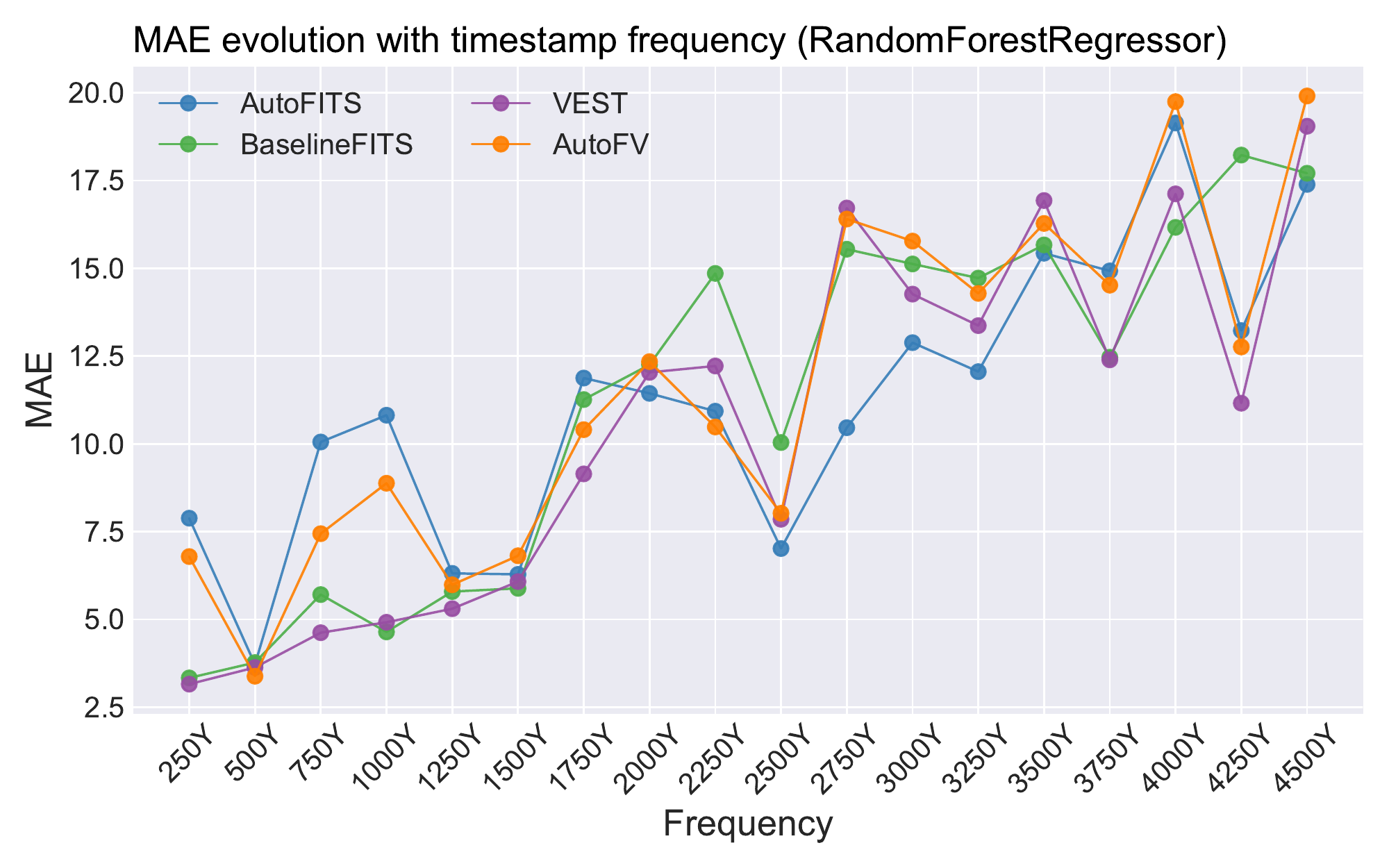}
    \caption{Vostok Ice Core dataset: MAE with a RandomForest learner.}
    \label{fig:mae_vostok_rf}
    \end{minipage}%
    \hspace{0.5cm}
    \begin{minipage}{.48\textwidth}
    \includegraphics[width=\linewidth]{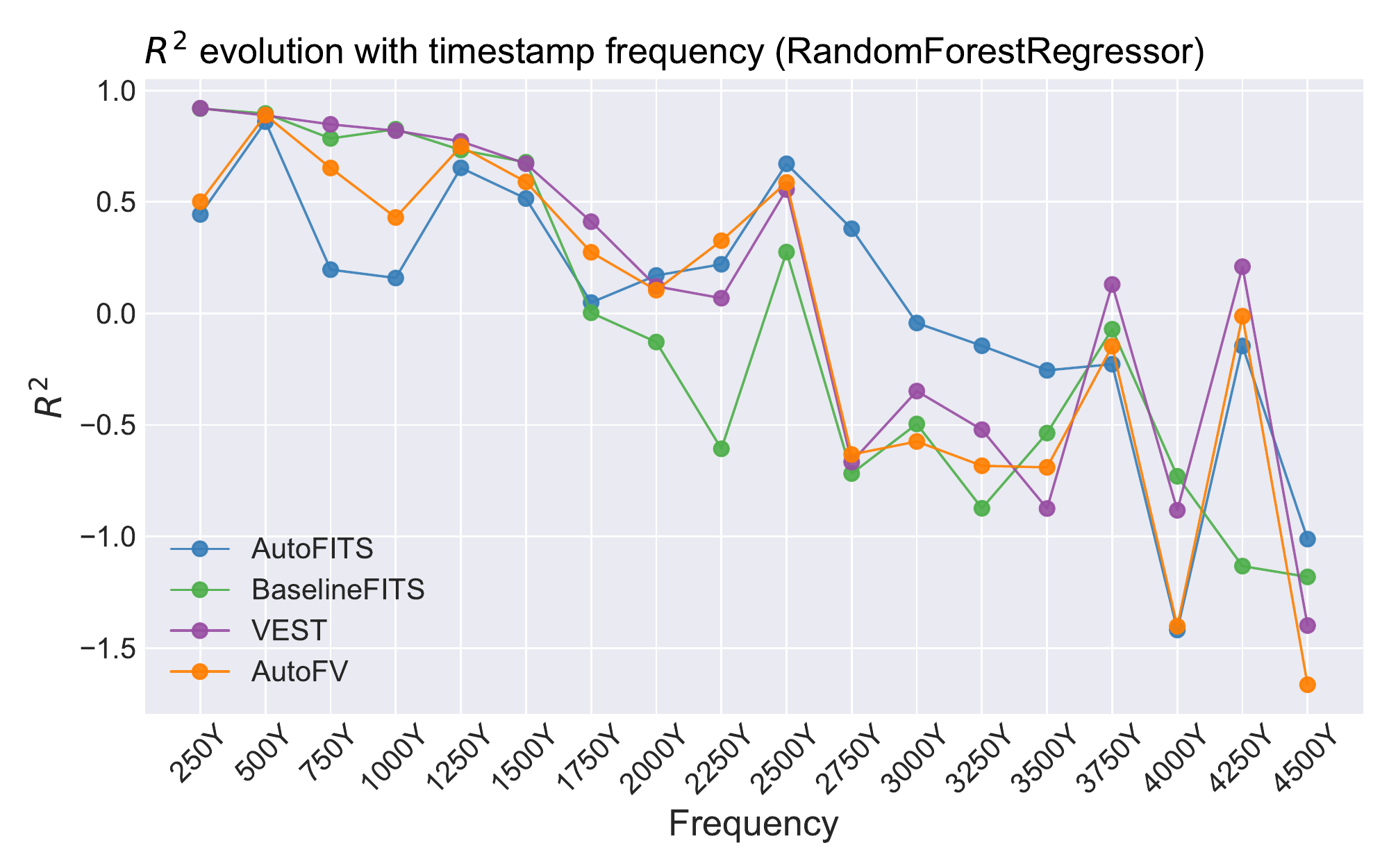}
    \caption{Vostok Ice Core dataset: \(R^2\) with a RandomForest learner.}
    \label{fig:r2_vostok_rf}
    \end{minipage}
\end{figure}

With a RandomForest learning algorithm, we get an overall increased MAE (Figure~\ref{fig:mae_vostok_rf}) and decreased \(R^2\) (Figure~\ref{fig:r2_vostok_rf}) and it appears that this algorithm might not be the best for this dataset when compared to LASSO.

In terms of \(R^2\), AutoFITS still outperformed all other models in 7 (39\%) cases, with VEST coming in second place by having the largest score in 6 (33\%) cases. However, if we consider only the cases where \(R^2 > 0\), we come to know that, for 5 different frequencies, no model manages to achieve a positive \(R^2\): 3000, 3250, 3500, 4000 and 4500. Curiously, in 4 of those, AutoFITS has the highest \(R^2\), so by rejecting negative scores, VEST becomes the clear winner by having the largest positive \(R^2\) in 6 out of 13 cases (46\%).

\subsubsection{Multiple entities}

\begin{figure}[htb]
    \centering
    \begin{minipage}{.48\textwidth}
    \includegraphics[width=\linewidth]{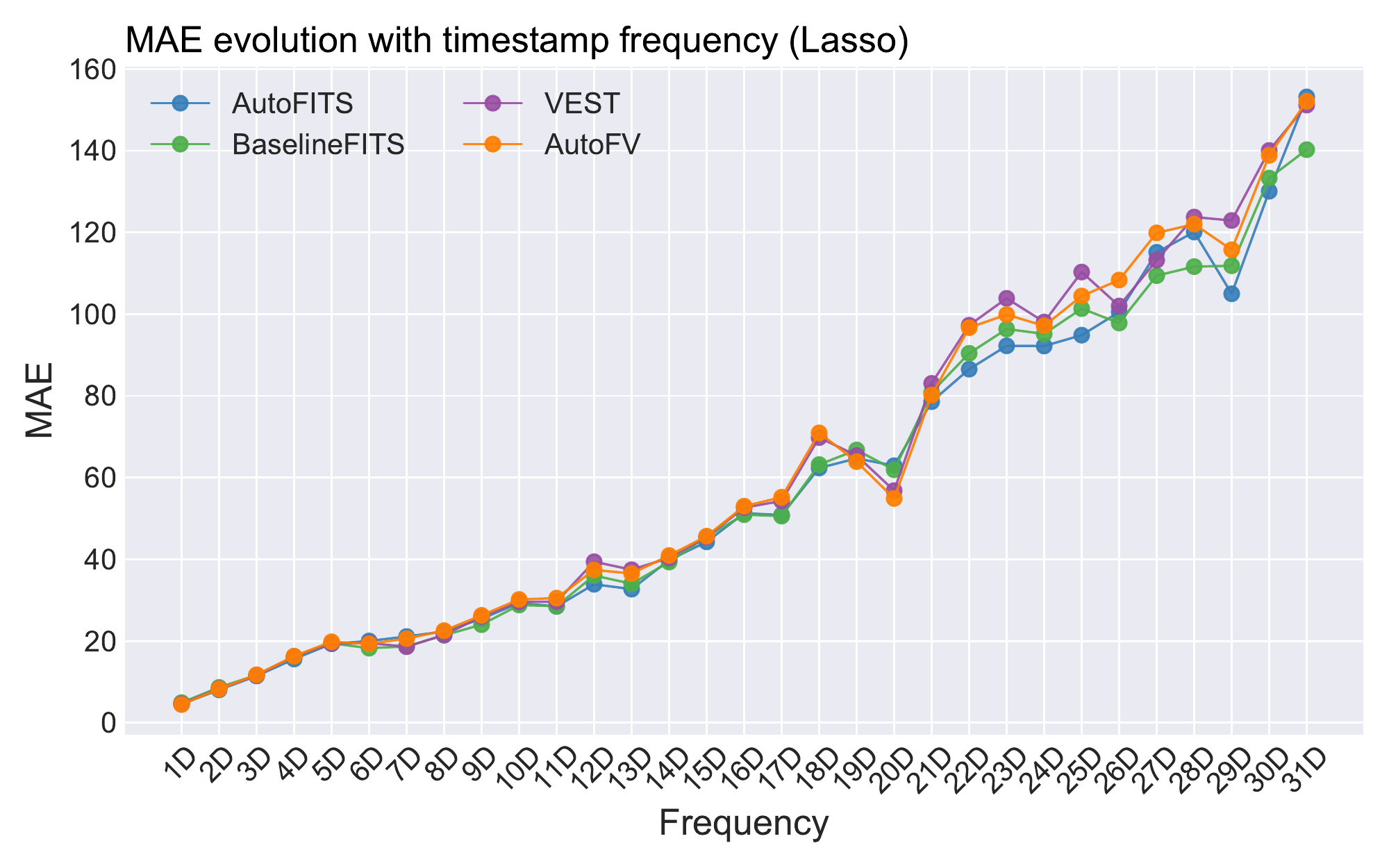}
    \caption{Recruit Restaurant Visitors dataset: MAE with a LASSO learner.}
    \label{fig:mae_rest_lasso}
    \end{minipage}%
    \hspace{.5cm}
    \begin{minipage}{.48\textwidth}
    \includegraphics[width=\linewidth]{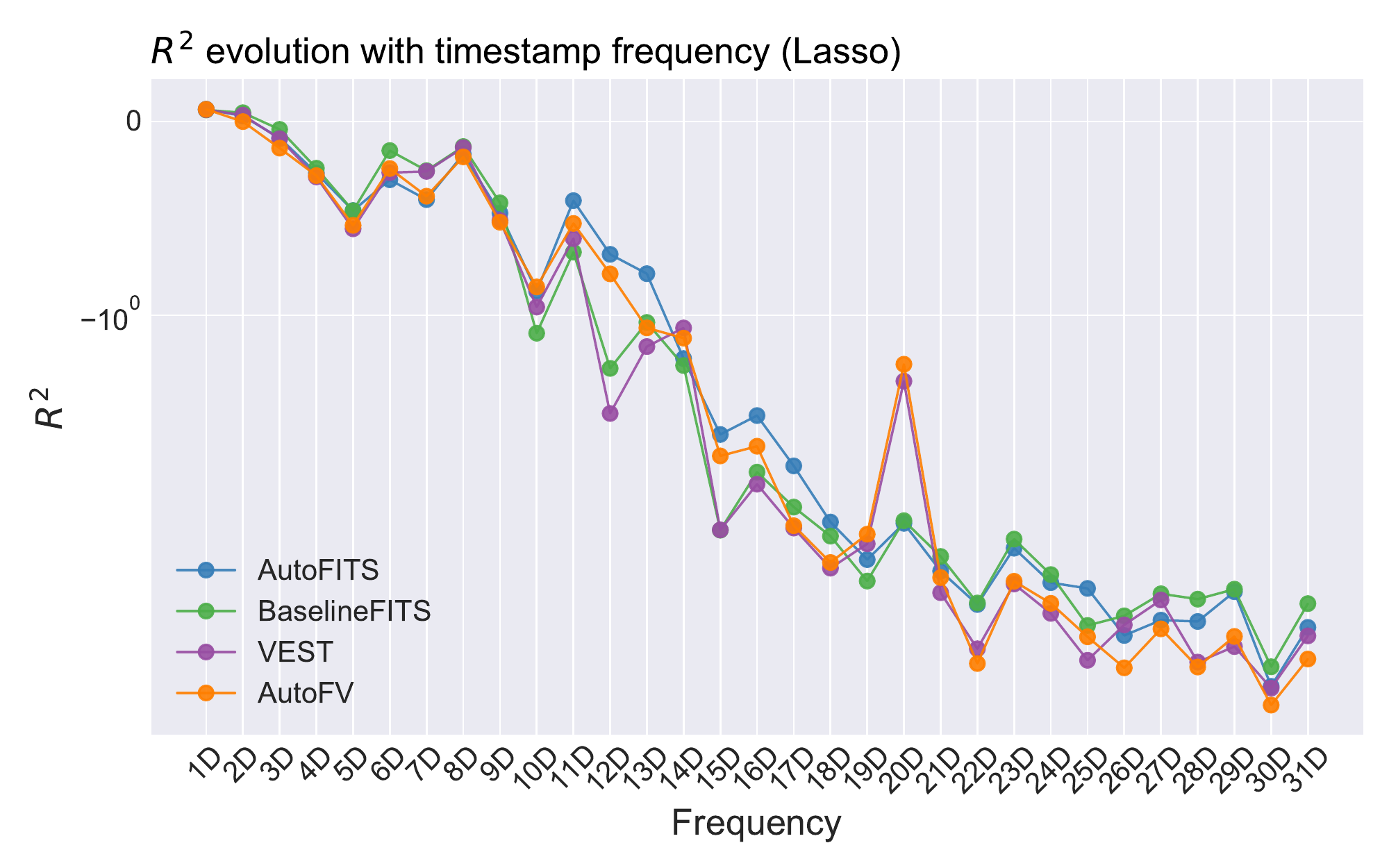}
    \caption{Recruit Restaurant Visitors dataset: \(R^2\) with a LASSO learner.}
    \label{fig:r2_rest_lasso}
    \end{minipage}
\end{figure}

In Figure~\ref{fig:mae_rest_lasso} we observe the obtained MAE for the Recruit Restaurant Visitors dataset with  a LASSO learning algorithm. It does not appear to exist a significant difference between the performance of all 4 models.

From the plot in Figure~\ref{fig:r2_rest_lasso}, we learn that all models were unable to achieve a positive \(R^2\) in all cases, except at frequencies 1 and 2. This, allied with what is seen in Figure~\ref{fig:mae_rest_lasso} allows us to conclude that LASSO does not seem to be indicated to handle this dataset.

\begin{figure}[htb]
    \centering
    \begin{minipage}{.48\textwidth}
    \includegraphics[width=\linewidth]{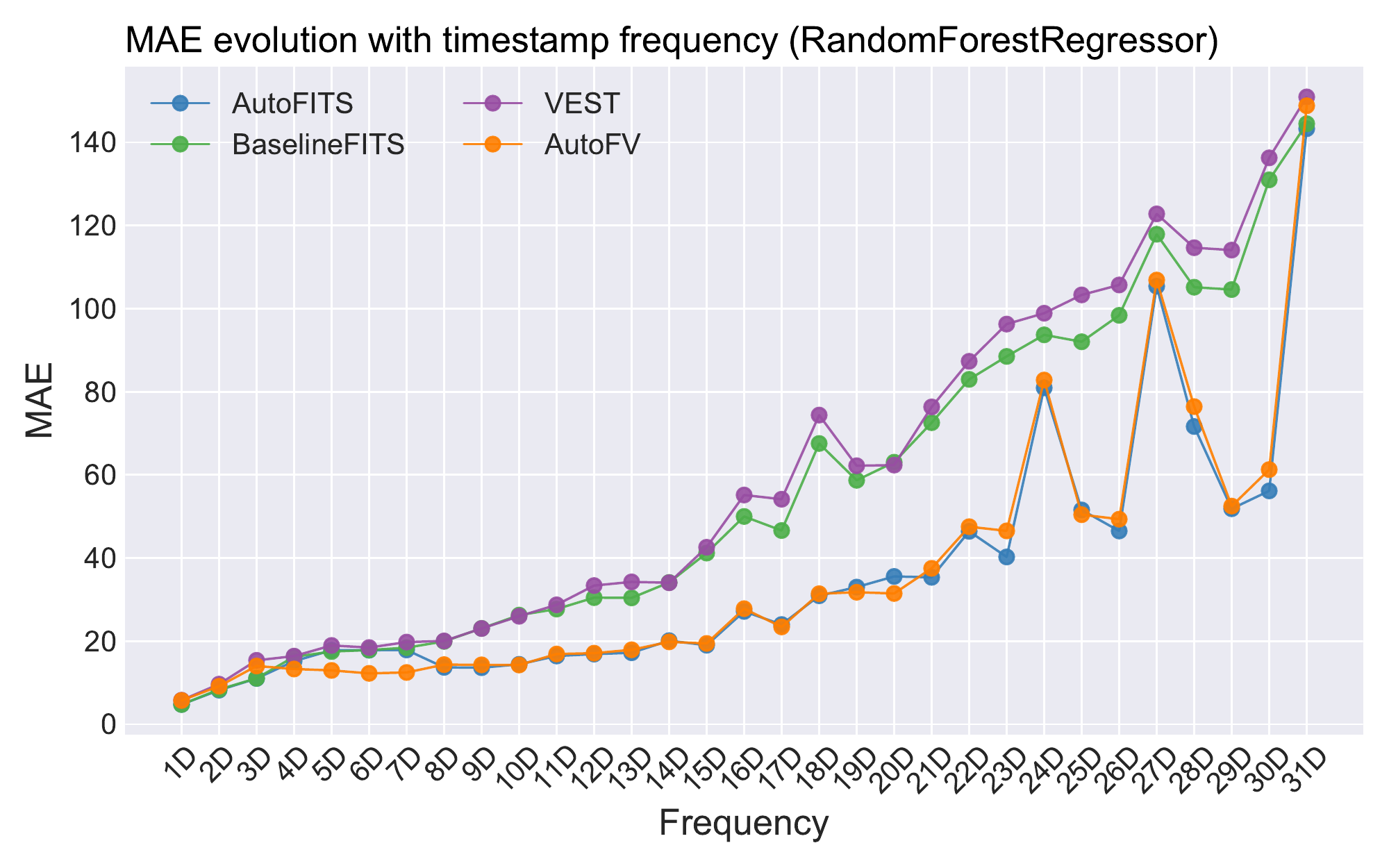}
    \caption{Recruit Restaurant Visitors dataset: MAE with a RandomForest learner.}
    \label{fig:mae_rest_rf}
    \end{minipage}%
    \hspace{.5cm}
    \begin{minipage}{.48\textwidth}
    \includegraphics[width=\linewidth]{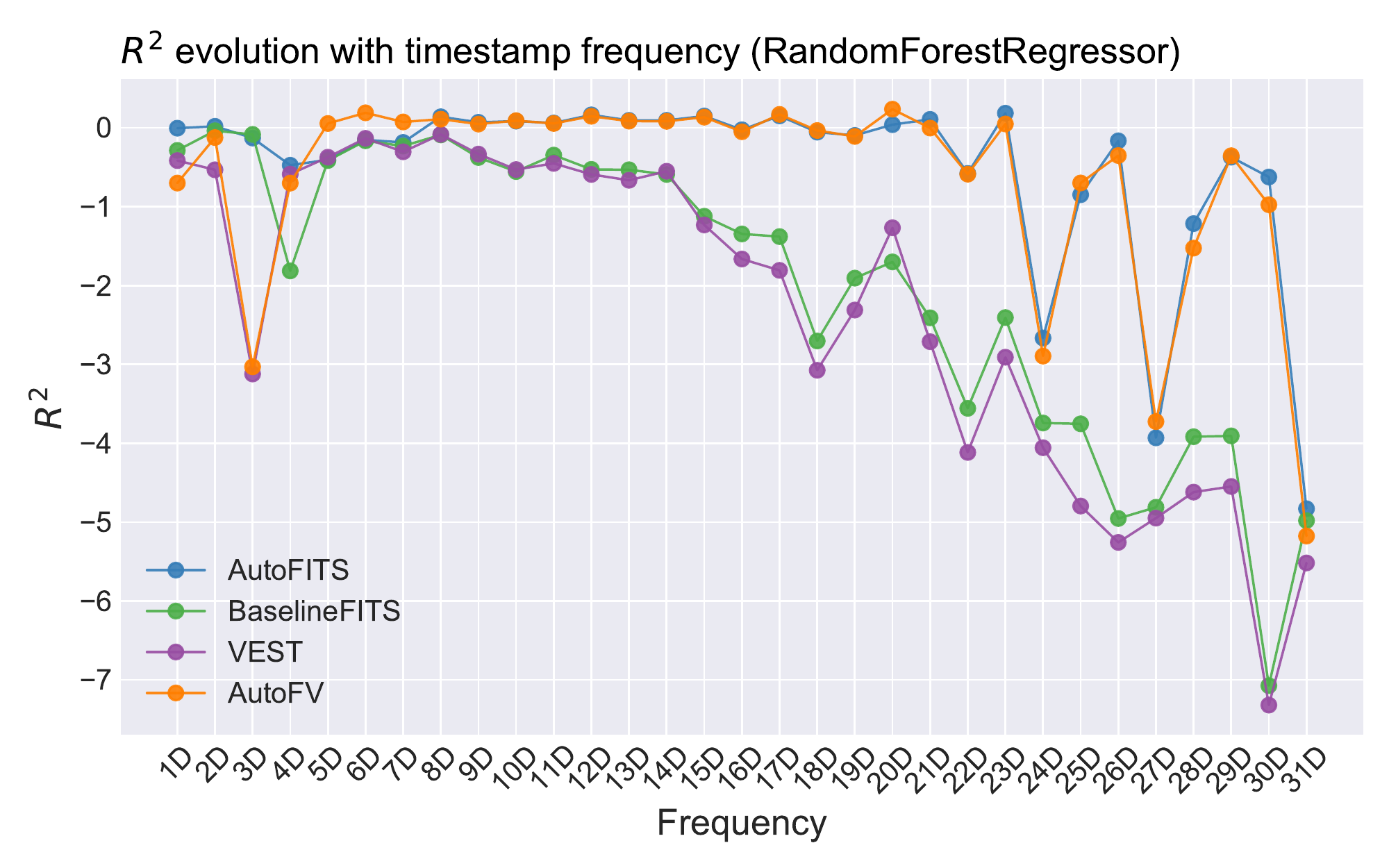}
    \caption{ Restaurant Visitors dataset: \(R^2\) with a RandomForest learner.}
    \label{fig:r2_rest_rf}
    \end{minipage}
\end{figure}

Contrary to Vostok, RandomForest seems to be the most adequate learner for the Recruit Restaurant Visitors dataset. In the MAE plot (Figure~\ref{fig:mae_rest_rf}), we see AutoFITS and AutoFV achieving significantly lower error then BaselineFITS and VEST. Relatively speaking, this is seen as a good indicator towards validating AutoFITS's value.

Also, with RandomForest we can have a positive \(R^2\) for many frequencies under 21 days with AutoFITS and AutoFV. However, \(R^2\) values very close to 0 suggests our learner is not being able to perceive most variability in the dataset. We would highlight the difference in performance between AutoFITS/AutoFV and BaselineFITS/VEST. There is a clear difference between the performance of both pairs of models in both MAE and \(R^2\). Although absolute values are far from ideal, results show that AutoFITS effectively improves over other approaches by exploiting irregularity.

\section{Conclusion}
\label{sec:conclusions}

In this work, we introduce AutoFITS: an automated feature engineering framework for irregular time series that aims to extract knowledge from irregularity. 
The objective of AutoFITS is to complement existing methods that do not extract features from irregularity. We studied how AutoFITS fares against a state-of-the-art method for time series feature engineering in the literature \cite{cerqueira2020vest}, and how it may help complement it. We evaluate our results under two perspectives: one where we study if explicitly modelling the irregularity of the time series provides useful information, and another where we evaluate if the built models can be applied in practice by generalising well to new observations.


Our results show that, by extracting features from irregularity, AutoFITS is able to improve over state-of-the-art feature extraction methods that do not account for irregularity, using both single-entity and multiple-entity time series.




\bibliographystyle{plain}
\bibliography{AutoFITS}

\end{document}